\documentclass[a4paper]{article}

\usepackage{amsmath}
\usepackage{amsfonts}
\usepackage{amssymb,bm}
\usepackage{pdflscape}
\usepackage[english]{babel}
\usepackage{babelbib}
\usepackage{dsfont}
\usepackage[utf8]{inputenc}
\usepackage[pdftex]{graphicx}
 \normalsize
\usepackage{fancyhdr}
\fancypagestyle{plain}{%
   \fancyhead{} 
}
\begin{document}
\pagestyle{empty}
\begin{center}
\begin{tabular}{c}
\end{tabular}
\vskip 60pt
{

  \Large Imitation of Life:
  \vskip 10pt
  \textbf{Advanced system for native Artificial Evolution}\\
  \normalsize
  
}
{

\vskip 40pt
}
{
  \large
  \textbf{by Sperl Thomas}\\
  \texttt{sperl.thomas@gmail.com}\\
  \vskip 10pt
  \texttt{July 2011}

\vskip 200pt
  \textbf{Abstract:}
}

\end{center}
A model for artificial evolution in native x86 Windows systems has been developed at the end of 2010. In this text, further improvements and additional analogies to natural microbiologic processes are presented. Several experiments indicate the capability of the system - and raise the question of possible countermeasures.

\newpage
\pagestyle{fancy}
\tableofcontents

\newpage
\section{Introduction}
Artificial Evolution has become a successful playground for evolutional experiments, when Tom Ray released the Tierra system.\cite{TomRay1}. Tierra is a virtual system with self-replicating programs which simulate mutations in form of copying errors. The artificial creatures struggle for the limited resources (such as CPU time and memory space), thus the systems fullfills the three criteria for evolution: replication, mutation, selection.\\

In order to achieve high robustness against mutations, Ray introduced techniques such as non-direct addressing and seperation of arguments and operations.\\

Many interesting insights to evolution have been found with Tierra (such as evolution of multi-cellularity\cite{TomRay2} or parallel computing\cite{TomRay3}) and similar systems such as avida (evolution under high mutation rate\cite{Lenski1} and emergence of complex features\cite{Lenski2}).\\

Iliopoulos, Adami and Sz\"or have discussed the consequences for computer security of implementing darwinian principles into native system, in 2008.\cite{Szor1} They concluded \emph{that a truly undetectable virus might be more feasible than previously imagined}, and that it is currently unknown whether there would be a defence against such organisms.\\

In 2010, I have created the first (to my knowledge) implementation of an artificial evolution system for a native operation system (Microsoft Windows XP+ 32bit), using several parallels to the natural biosynthesis process\cite{Thomas1}. A short comparison between usual x86 code and the new artificial evolution concept shows that the new concept is actually more robust against mutations.\cite{Thomas2}\\

The presented proof-of-concept organism as well as the underlaying metalanguage have been analysed in detail by Peter Ferrie in 2011\cite{PF1}\cite{PF2}.\\

The idea of this article is to continue this research...

\section{Artificial Biosynthesis}
The main idea is to use a similar concept to natural biosynthesis: The codons of the mRNA are translated into amino acids using tRNA molecules, these amino acid chains form the proteins - the actually functional part of the cell.\\

In the artificial analogon, the whole information of the organism is saved in a chain of codons. Each codon consists of 8bits, thus there are 256 different codons. In the translator (similar to tRNA in the ribosom), the codons will be mapped to an x86 instruction (similar to an amino acid) - a chain of x86 instructions form the protein, the functional part of the organism:\\

\begin{center}  
\begin{tabular}{ c | c}
 \textbf{Artificial}   &   \textbf{Natural}\\
 bit          &   nucleobase \\
 byte         &   codon\\
 instruction  &   amino acid\\
 function     &   protein\\
 translator   &   tRNA
\end{tabular}
\end{center}

In the natural system, the information is saved in codons, which consists of 3 nucleobases - as there exists 4 different nucleobases (\textbf{A}denine, \textbf{G}uanine, \textbf{T}hymine, \textbf{C}ytosine), a codon can have $4^3$ = 64 different representations. Each codon codes one out of 20 amino acids, thus there is a redundancy in the mapping process, which is used to increase the robustness of the code. This redundancy is also used in the artificial biosynthesis as there are less than $2^8$=256 base functions of the meta language.\\

\subsection{Meta-Language}
The idea is to create a compact, \emph{complete} instruction set with seperation of arguments and operations, and with non-direct addressing.\\

The language provides seven registers with specific properties:\\

\begin{itemize}
\item \textbf{RegA, RegB, RegD} - general purpose registers (correspond to EAX, EBX, EDX)
\item \textbf{BC1} - operation register (correspond to EBX): the first argument of every operation, and source or destination for other instructions 
\item \textbf{BC2} - argument register (correspond to ECX): the second argument of every operation
\item \textbf{BA1} - write address register (correspond to EDI): holds the address for write instructions
\item \textbf{BA2} - jump address register (correspond to ESI): holds the address for jump instructions     
\end{itemize}

The seperation of arguments and operations is realized by using BC1 (and BC2) as standard arguments, and filling them independently of the operation.\\

The language provides PIC (\textit{position-independent code}). Every address is relative to the instruction pointer thus is independent of the position.\\

\begin{center}  
\begin{tabular}{ l | l | l}
\textbf{\begin{Large}Instruction\end{Large}} & \begin{Large}HLL\end{Large} & \begin{Large}Assembler\end{Large} \\ \hline \hline \hline
\textbf{nopREAL} &  & nop \\ \hline \hline
\textbf{nopsA} & BC1 = RegA & mov ebx, eax \\ \hline
\textbf{nopsB} & BC1 = RegB & mov ebx, ebp \\ \hline
\textbf{nopsD} & BC1 = RegD & mov ebx, edx \\ \hline
\textbf{nopdA} & RegA = BC1 & mov eax, ebx \\ \hline
\textbf{nopdB} & RegB = BC1 & mov ebp, ebx \\ \hline
\textbf{nopdD} & RegD = BC1 & mov edx, ebx \\ \hline \hline
\textbf{save} & BC2 = BC1 & mov ecx, ebx \\ \hline
\textbf{addsaved} & BC1 + = BC2 & add ebx, ecx\\ \hline
\textbf{subsaved} & BC1 - = BC2 & sub ebx, ecx\\ \hline \hline
\textbf{saveWrtOff} & BA1 = BC1 & mov edi, ebx \\ \hline
\textbf{saveJmpOff} & BA2 = BC1 & mov esi, ebx \\ \hline \hline
\textbf{writeByte} & byte[BA1] = (BC1 \& 0xFF) & mov byte[edi], bl \\ \hline
\textbf{writeDWord} & dword[BA1] = BC1 & mov dword[edi], ebx \\ \hline \hline
\textbf{getDO} & BC1 = DataOffset & mov ebx, DataOffset \\ \hline
\textbf{getdata} & BC1 = dword[BC1] & mov ebx, dword[ebx] \\ \hline
\textbf{getEIP} & BC1 = InstructionPointer & call gEIP; gEIP: pop ebx \\ \hline \hline
\textbf{push} & push BC1 & push ebx \\ \hline
\textbf{pop} & pop BC1 & pop ebx \\ \hline
\textbf{pushall} & pushad & pushad \\ \hline
\textbf{popall} & popad & popad \\ \hline \hline
\textbf{zer0} & BC1 = 0 & mov ebx, 0x0 \\ \hline
\textbf{add0001} & BC1 + = 0x1 & add ebx, 0x1 \\ \hline
\textbf{add0004} & BC1 + = 0x4 & add ebx, 0x4 \\ \hline
\textbf{add0010} & BC1 + = 0x10 & add ebx, 0x10 \\ \hline
\textbf{add0040} & BC1 + = 0x40 & add ebx, 0x40 \\ \hline
\textbf{add0100} & BC1 + = 0x100 & add ebx, 0x100 \\ \hline
\textbf{add0400} & BC1 + = 0x400 & add ebx, 0x400 \\ \hline
\textbf{add1000} & BC1 + = 0x1000 & add ebx, 0x1000 \\ \hline
\textbf{add4000} & BC1 + = 0x4000 & add ebx, 0x4000 \\ \hline
\textbf{sub0001} & BC1 - = 0x1 & sub ebx, 0x1 \\ \hline \hline
\textbf{shl} & BC1 $<<$ (BC2 \& 0xFF) & shl ebx, cl \\ \hline
\textbf{shr} & BC1 $>>$ (BC2 \& 0xFF) & shr ebx, cl \\ \hline
\textbf{xor} & BC1 $\wedge$ = BC2 & xor ebx, ecx \\ \hline
\textbf{and} & BC1 \& = BC2 & and ebx, ecx \\ \hline \hline
\textbf{mul} & (RegD: RegA) = RegA * BC1 & mul ebx \\ \hline
\textbf{div} & (RegD, RegA) = RegA / BC1 & div ebx \\ \hline \hline
\textbf{JnzUp} & & jz over; jmp esi; over: \\ \hline
\textbf{JnzDown} & & jnz down; times 32: nop; down: \\ \hline
\textbf{call} & & stdcall ebx \\  \hline
\textbf{CallAPILoadLibrary} & & stdcall dword[LoadLibrary]
\end{tabular}
\end{center}
\subsubsection{Mutable API calls}

To use the APIs provided by the OS, an organism can use the \texttt{call} instruction:\\

\begin{center}
\begin{tabular}{c}
  \includegraphics[scale=0.73]{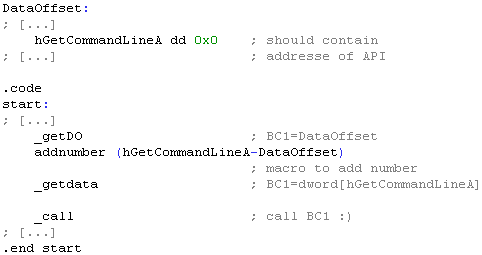}
  \end{tabular}
\end{center}

One possibility is to save the API addresses directly within the organism body. However, this has two disadvantages: Firstly, the addresses may change for each new version of Windows. Secondly, the probability that a different and valid API address appears out of one single bitflip is very low (a rough estimate: a memory address of 32-bit gives $2^{32}$ possible addresses - say there are 5000 valid API addresses. The probability to reach one of them is $P = \frac{5000}{2^{32}} \approx 10^{-6}$).\\

A different method is to save a short hash of the desired API name in the organism. Then load the DLL, scan the export section for API names and create hashes for each API. If the hashes match, save the address of the API. This approach is independent of the OS version, and it's very mutable. It is possible to use hash as short as 12bit for each API. Let's say there are 1000 APIs in a DLL file. The probability to access on of these APIs within one single bitflip is given by $P = \frac{1000}{2^{12}} \approx 0.25$. In average, every 4th bitflip in the API hash leads to a different hash corresponding to a valid API.\\

\subsubsection{Example: ROL instruction}
The x86 instruction ROL (\textit{Rotate Left}) is not directly provided by the instruction set. However, it can be written using just instructions of the set.\\

Let's say, one would like to write \texttt{rol RegA, c}, where c is some integer - then the usual assembler instructions looks like this:\\

\includegraphics[scale=0.73]{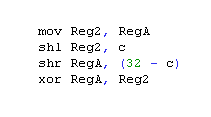}\\

Translating this into the ArtEvol MetaLanguage is very easy:\\

\includegraphics[scale=0.73]{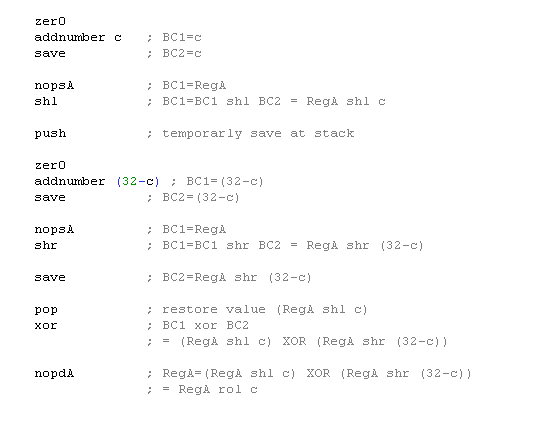}\\

The 2nd argument (\texttt{c} in this case) is saved in the BC2 register, then the first argument is loaded into the BC1 register and the operation is performed. As ROL requires three operations, the result of the first one is temporarily saved at the stack; the same procedure is performed again, and in the end, the results are combined.\\

The \texttt{addnumber} is a macro which returns the right combination of \texttt{addNNNN} instructions.\\

\subsection{Translator}
In order to convert the metalanguage instructions to native x86 instructions, a tiny translator is used:\\

\begin{center}
\begin{tabular}{c}
  \includegraphics[scale=0.73]{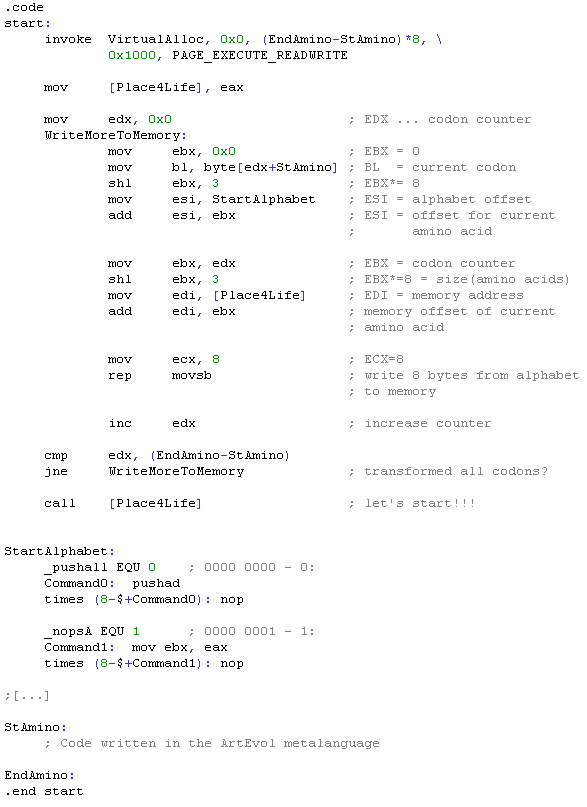}
  \end{tabular}
\end{center}

\subsection{Replication, Mutation and Selection}
To achieve Evolution, a system requires Reproduction, Mutations and Selection.\\

In the \textbf{reproduction} stage, the organism creates a living copy of itself. In the artificial system of a computer, this can be done in the same way as computer viruses and computer worms do - interfere with special file formats or network protocols such that the copy will be executed in a different habitat (other computer, other file, ...). However, this requires alot of previous knowledge about the system, thus is not the simplest starting point for evolution. The most trivial way of reproduction is to copy the own file in the current directory and run it - which is actually the way how it is done in the experiments.\\ 

The interference between reproduction and \textbf{mutations} leads to non-identical replica of the organism. In the biological system, mutations happens due to disruptive effect such as cosmic X-rays. This leads to point mutations (exchange of one single codon) or more difficult chromosome abnormality. The natural mutation probability in a computer system (such as mistakes in the copy process) is neglectable, thus the organism has to carry its own mutation engines. This is explained in detail in the next chapter.\\

Natural \textbf{selection} is a process in which a certain trait becomes more or less common in the population, depending on its effect on the fitness of the organism. This process appears when the organism compete, struggle for limited values (such as energy), or are exposed to natural enemies. In the artificial system of a computer, anti virus programs could be responsible for natural selection (and by that unwillingly initiate a faster evolutionary process at all). A different selective pressure comes from attentive user, who would stop any suspicious behaviour.\\

\section{Mutations}
\subsection{Point mutation - bit flip}
Point mutations change single nucleobases in the DNA. These mutations can be categorized into silent (when the affected codon maps to the same amino acid due to the redundance in the alphabet), missense (when the affected codon maps to a different amino acid); or nonsense (when the affected codon maps to the STOP codon).\\

A native analogon to that concept would be the change of single bits in the organism - called \textit{Bitflip}. These mutations have the same categories as their biological companion.\\

The mutation rate (mutations per base per generation) in biological organisms varies from $10^{-4}$ for very small (some kilo bases) to $10^{-8}-10^{-9}$ for humans (some giga bases). Finding an adequate mutation rate for artificial organisms is not trivial, as too small values lead to mainly unmutated offspring, thus no evolution; whereas too big mutation rates lead to extinction of the population. To get the best, one could test organisms with different mutation rate, and take the critical value.\\

\begin{center}
\begin{tabular}{cc}
\includegraphics[scale=0.6]{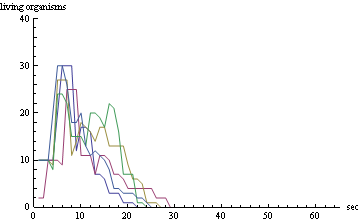} & \includegraphics[scale=0.6]{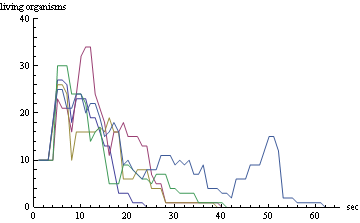}\\
\textbf{Figure 3.1}: Five populations with a & \textbf{Figure 3.2}: Five populations with a \\
mutation rate of 1 / 7.001 &  mutation rate of 1 / 9.001\\
\end{tabular}
\end{center}

\begin{center}
\begin{tabular}{cc}
\includegraphics[scale=0.6]{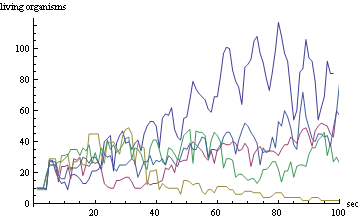} & \includegraphics[scale=0.6]{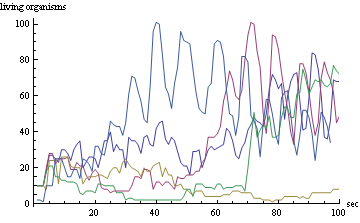}\\
\textbf{Figure 3.3}: Five populations with a & \textbf{Figure 3.4}: Five populations with a \\
 mutation rate of 1 / 11.003 & mutation rate of 1 / 13.499\\
\end{tabular}
\end{center}

The test-organism - which can create three offspring - has a size of 20.480 bytes, the critical mutation rate is between $\frac{1}{9.001}$ and $\frac{1}{11.003}$; the probability that at least one bit is changed is between 84.5\% and 89.7\%.\\   

\subsection{Chromosomal inversion - byte eXCHanGe}
A different kind of mutation happens when a segment of a chromosome is reversed. This happens when a segment breaks off and is rearranged in the wrong way.\\

A similar method is used in this project: two consecutive d-words (that means, 4 codons each) are exchanged. This mutation is not as dangerous as it may look like at first; the codon streams forming a functional part are very big, thus are not too sensitive on such local translocations.\\

\begin{center}
\begin{tabular}{cc}
\includegraphics[scale=0.6]{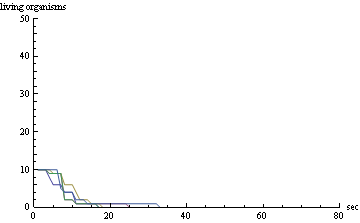} & \includegraphics[scale=0.6]{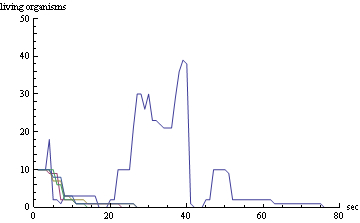}\\
\textbf{Figure 3.5}: Five populations with a & \textbf{Figure 3.6}: Five populations with a \\
 mutation rate of 1 / 26.501 & mutation rate of 1 / 27.011\\
\end{tabular}
\end{center}

\begin{center}
\begin{tabular}{cc}
\includegraphics[scale=0.6]{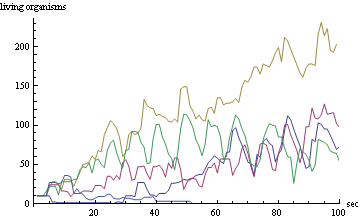} & \includegraphics[scale=0.6]{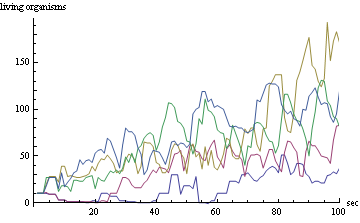}\\
\textbf{Figure 3.7}: Five populations with a & \textbf{Figure 3.8}: Five populations with a \\
 mutation rate of 1 / 27.253 & mutation rate of 1 / 27.509\\
\end{tabular}
\end{center}

These graphs show that the critical mutation rate is much sharper than for Bitflips - this is what you would expect as Byte Exchange has much stronger effects.\\

\subsection{Deletion, insertion, translocation}
Deletion is a mutation in which a part of the DNA is missing; insertion is the inverse process, where an additional sequence of DNA is included into the genom. Translocation is a combination of these mutations: A part of the DNA breaks off and is included at a different place.\\

A similar method can be realized in artificial organisms in the computer system within one simple algorithm. Three random values are calculated (the place of the mutation $P$, the size of the inserted NOP block $S_i$ and the size of the translocated block $S_b$). Then at P, a block with the size of $S_b$ will be translocated $S_i$ bytes, the new created block at P will be filled with NOPs.\\
 
\newpage 
\begin{center}
\includegraphics[scale=0.322]{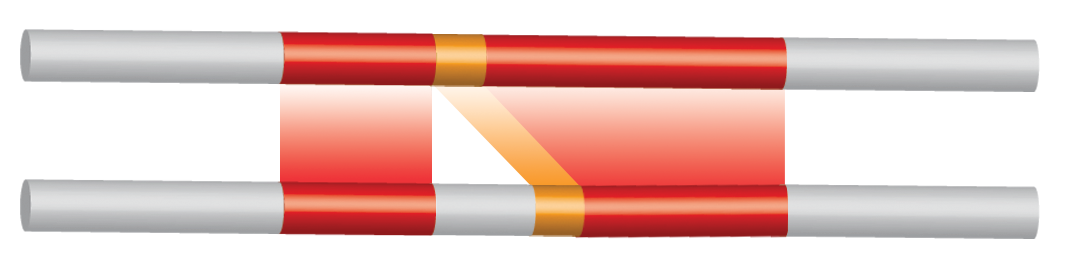}\\
\end{center}
\textbf{Figure 3.9}: Deletion process in the organism. Grey are NOPs, red and orange are functional parts. Red is not moved, orange is the translocated sequence. The second red part is smaller after deletion, as it has lost some of its code.\\

\begin{center}
\includegraphics[scale=1.4]{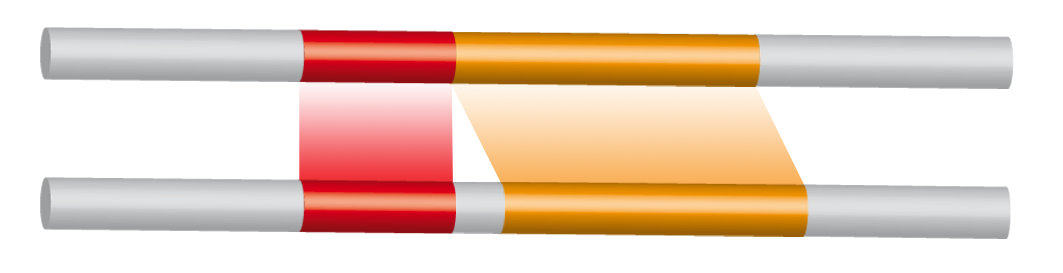}\\
\end{center}
\textbf{Figure 3.10}: Insertion process in the organism. Grey are NOPs, red and orange are functional parts. Red is not moved, orange is the translocated sequence. The whole orange part is translocated, thus there is nothing deleted - just an insertion of NOPs.\\

\subsection{Horizontal gene transfer}
Horizontal gene transfer is a process in which an organism incorporates genetic material from another organism without being the offspring of that organism. In biological systems, this is a controlled (not by random mutations) method to receive beneficial functions such as antibiotic resistance. Photosynthesis is an important process which has been developed with horizontal gene transfer from different bacteria.\\

In the artificial system, an organism could try to interact with other organism and exchange valid code, and therefore perform a symbiotic \textit{conjugation}. However, this would require a specific protocol for communication (such as F-plasmids in bacteria) - which is not developed so far.\\

Nevertheless the artificial organism could try to gain new information from other files, just by opening them and copy some parts of their code. In the case that the other file is written in the same language, the organism has the chance of getting new functions.\\

\subsection{*Polymorphism: neutral codon variation}
In the artificial organisms, the alphabet has $2^8 = 256$ entries which map to 45 or less instructions, thus there is a big redundancy - that means several codons map to the same amino acid.\\

The organism can scan thru its alphabet, detect equal amino acids, then scan its codon-stream and exchange the codons which point to the same amino acid.\\

There are a few advantages to use this technique in artificial organism:
Firstly, codons which point to isolated amino acids (these who can not be transformed by a single bitflip to another amino acid of the same kind) can be de-isolated, thereby increase the robustness of the overall code. Secondly, such \textit{macro mutations} are of high importance to bypass natural enemies (such as antivirus software), thus increase fitness. And thirdly, mutations in the polymorphism-engine itself or variations of the START or STOP codon could lead to unpredictable results.\\

\section{Further improvements}
\subsection{Start- and Stop codons: Splicing}
Natural genetic code contains alot of non-functional garbage, which can be old unused DNA or (malformed) duplicates of actual functional code. In human DNA, approximately 95\% of the DNA is garbage. These non-coding parts are called \textit{Introns}, the functional parts are \textit{Exons}. Before translation of mRNA into proteins, the introns are cutted out in a process called \textit{Splicing} - this is done by taking usage of two special codons - the \textit{START} and the \textit{STOP} codon. Each functional part starts with a START-codon and ends with a STOP-codon - the parts between a STOP codon and the next START codon (which is actually the intron) will be removed.\\

The advantage of such introns is that within the unused code new DNA sequences could be developed which are (in very rare cases) actually functional - and have different behaviour. Together with the additional possibility for altering its code, this would be a special gain for artificial organisms.
\begin{center}
\begin{tabular}{c}
  \includegraphics[scale=0.73]{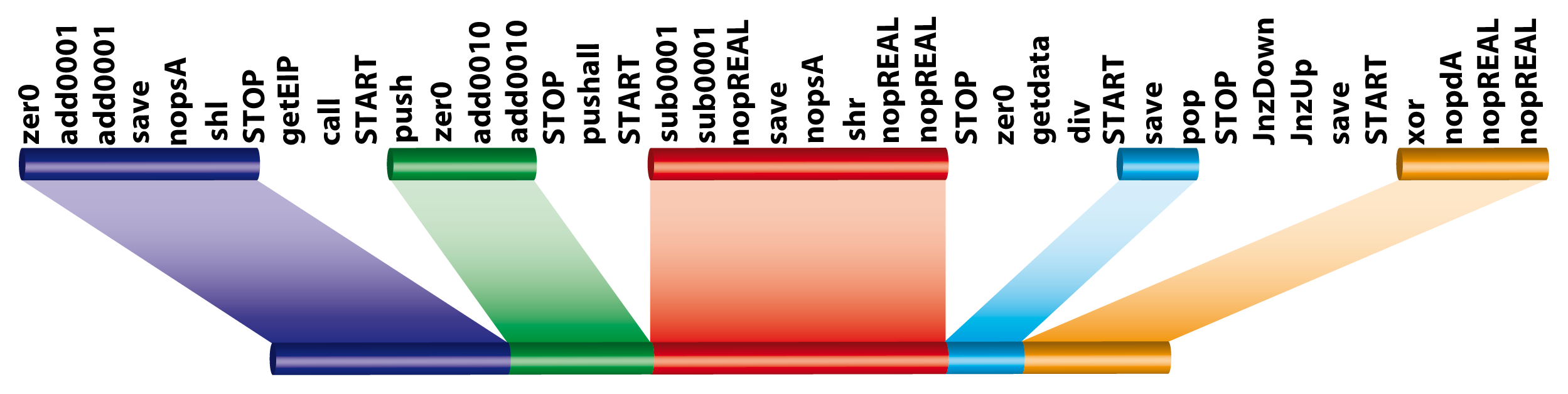}
  \end{tabular}
  \textbf{Figure 4.1}: Splicing functionality in a artificial organism
\end{center}

Implementing a very small splicing algorithm into the translator can be achieved in the following way:\\

\begin{itemize}
\item All codons in the form of \textbf{1??1.???1} (32 codons) point to a NOP amino acid
\item At the translation, whenever there is a STOP codon, AL=0x91 (1001.0001); whenever there is a START codon, AL=0x0
\item Each codon will be OR'ed with AL
\item In the end, each codon after a STOP mark will be redirected to a NOP amino acid - until there is a START codon
\end{itemize}

\begin{center}
\begin{tabular}{c}
  \includegraphics[scale=0.73]{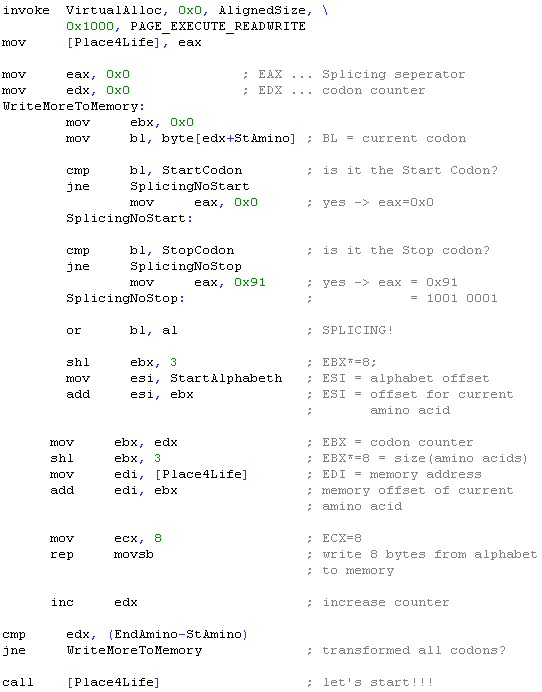}
  \end{tabular}
\end{center}

\subsection{Optimization of the instruction set}
The original instruction set had 43 different entries. Ofria, Adami and Collier found out, that a smaller instruction set leads to higher robustness under mutations, thus higher fitness\cite{Ofria1}.\\

In their experiments, the reason is that small instruction set requires bigger realisations of function, thus the risk of a lethal mutation is spread over a larger area. In our realisation, a second advantage appeares. A small instruction set leads to a more redundant alphabet, therefore allows more codons to point to the same amino acid. In the end, there is a bigger probability that a single BitFlip changes the codon such that it still points to the original amino acid.\\

Peter Ferrie was able to create an optimized instruction set with just 18 entries, by replacing an instruction with a combination of other instructions.\cite{PF2} One simple example is \texttt{zer0} $\to$ \texttt{save}+\texttt{xor}.\\

For implementing these optimizations, one has to take care of changed Registers and Flags. This can be done by using Stack instruction (\texttt{pushall} and \texttt{popall}) or even implement a new \textit{pseudo}register (in the .data section) with some special properties.\\

Unfortunately, these implementations lead to an excessive useage of \textit{evolutionary dangerous} instructions, which are instructions that lead to the programs crash if they are replaced by other instructions.\\

I consider dangerous instructions as everything that interacts with the stack (\texttt{push}, \texttt{pop}, \texttt{pushall}, \texttt{popall}, \texttt{CallAPILoadLibrary}), that influences the code flow (\texttt{JnzDown}, \texttt{JnzUp}, \texttt{call}, \texttt{saveJmpOff}), and that interacts with the memory (\texttt{saveWrtOff}, \texttt{writeByte}, \texttt{writeDWord}, \texttt{getdata}). One can create two further categories: \textit{semi-harmless}, which are all instructions that change the values of \texttt{RegA}, \texttt{RegB}, \texttt{RegD} and \texttt{BC2}, and \textit{harmless} instructions just change the \texttt{BC1} register.\\

\begin{center}
\begin{tabular}{cc}
\includegraphics[scale=0.6]{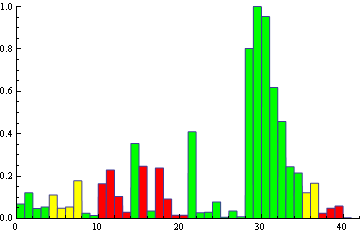} & \includegraphics[scale=0.6]{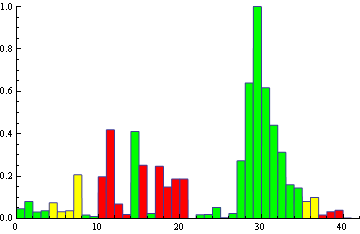}\\
\textbf{Figure 4.2}: Original instruction set & \textbf{Figure 4.3}: \texttt{zer0} removed \\
\end{tabular}
\end{center}

\begin{center}
\begin{tabular}{cc}
\includegraphics[scale=0.6]{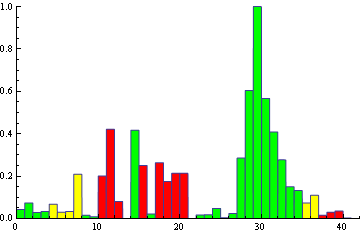} & \includegraphics[scale=0.6]{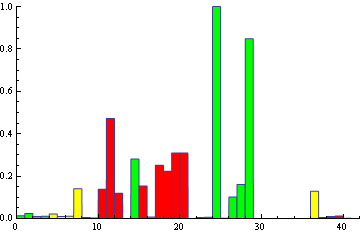}\\
\textbf{Figure 4.4}: \texttt{subsaved} removed & \textbf{Figure 4.5}: \texttt{addNNNN} removed \\
\end{tabular}
\end{center}

\begin{center}
\begin{tabular}{cc}
\includegraphics[scale=0.6]{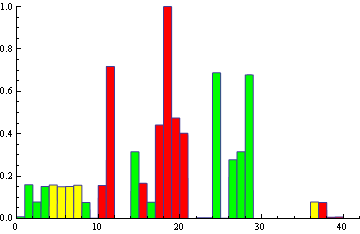} & \includegraphics[scale=0.6]{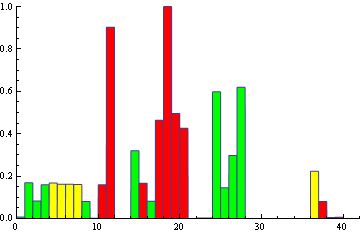}\\
\textbf{Figure 4.6}: \texttt{addsaved} removed & \textbf{Figure 4.7}: \texttt{add0001} removed \\
\end{tabular}
\end{center}

The x-axis shows the different instructions (with the same order as written in chapter 2), the y-axis gives the (normalized) appearence of the instruction in an organism. It is interesting to see how dangerous instruction density increases, especially after removing the \texttt{addsaved} instruction.\\

\subsection{Optimization of the alphabet}

To achieve the optimal robustness, evolution has lead to a special order of codon mapping. As an example, the amino acid \textit{Proline} can be coded via \texttt{CCU}, \texttt{CCC}, \texttt{CCA} and \texttt{CCG}. That means, whenever a mutation changes the third nucleobase, still a codon remains that codes Prolin (in biological systems, mutations happens most often at the third nucleobase).\\

For artificial organisms, one has 256 slots for about 45 instructions, and furthermore different pairs of exchanged codons have different probability to cause mistakes (exchanging \texttt{add0001} with \texttt{add0004} may cause less problems than exchanging \texttt{add0001} with \texttt{pushall}).\\

This is a nonlinear problem, and solutions by hand take long and are of low quality. However, one can reformulate the problem - with the help of a bit of physics:\\

Let's imagine the codons as objects in a special space, such that each two codons with one bit difference are neighbors (the geometry of this space is an 8 dimensional cube with codons on the corners). For example, the codon \texttt{0000.0000} and \texttt{0010.0000} are neighbors. Each codon interacts with its neighbors, thus has an interaction energy (which depends on the types of the codons).\\

We can define V(A,B) as the interaction energy of codon A and B. One possible definition would be\\

\begin{tabular}{ l l }
  V(A,B) = 0 & if A = B \\
  V(A,B) = 0.5 & if A and B are \texttt{addNNNN} or \texttt{sub0001} \\
  V(A,B) = 0.66 & if A and B are harmless instructions \\
  V(A,B) = 0.75 & if A and B are harmless or semi-harmless instructions\\
  V(A,B) = 1 & else (if A or B is a dangerous instruction or a START or STOP codon)\\
\end{tabular}

Now we can define the total energy of the system as:
\begin{align}
E_{\textnormal{total}}=\sum_{i=1}^{256} \sum_{j=1}^8 V(\textnormal{codon}_i, \textnormal{codon}_j)\nonumber
\end{align}
The total energy of a maximum random system would be 2.048, the mimumum total energy of a system with just one single instruction would be zero.\\

Finally, we can reformulate "find an optimal alphabet" into "find the minimal interaction energy of the system". There are several ways to find a minimum energy, one is the Metropolis-Algorithm; we use a slightly modified one.\\

First, we fill the 256 entries with instructions of random order, and calculate the total energy of the system. Then we exchange a few instructions randomly, and calculate the new total energy. If the new energy is smaller than the old one, we keep the new system, otherwise we continue with the old one. To find a (local) minimum, one can repeat that method.\\

\begin{center}
\includegraphics[scale=0.8]{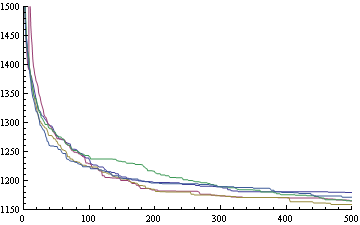}\\
\end{center}
\textbf{Figure 4.8}: Energy minimization of five systems. x-axis is iteration number (in 1000s), y-axis is energy.

The figure shows that the alphabet reaches a good local minimum after a few 100.000 iterations. It also shows that systems with different starting order find local minimal of approximately the same energy.\\

\section{Experiments}
An experiment measures the fitness of the organisms by letting them struggle for limited resources (such as CPU time and memory). To control the experiment, several \textit{guard files} ran in the background, to close endless-loop files, multiple instances of the same file, unmutated files by a certain probability. A more detailed explanation can be found in \cite{Thomas1}.
\subsection{Hamming distance}
In the first experiment, we analyse the \textit{long time} behaviour of a population, that can just perform BitFlips and Byte XCHG.\\

The Hamming distance (difference in the bit-code) with respect to the original ancestor is calculated every 3 minutes.\\

\begin{center}
\includegraphics[scale=0.8]{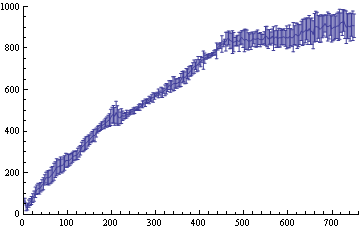}\\
\end{center}
\textbf{Figure 5.1}: Evolution of Hamming distance over 12 hours. x-axis: Hamming distance, y-axis: time (in minutes)\\

It is very surprising and interesting, that after about 7.5 hours a mutation has occured with the effect that some organisms can largely bypass the no-cloning guard. A deeper analysis of that event would require reverse engineering of mutated code, which is a non-trivial task - and therefore hasn't been done yet.\\

However, this event is an indication that artificial organisms can bypass control instances very fast.\\

A second experiment has been performed, with an organism which contains about 90\% introns (similar to natural organisms).\\
\begin{center}
\includegraphics[scale=0.9]{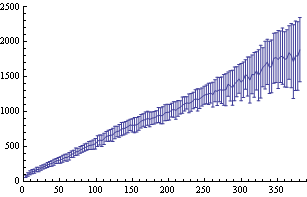}\\
\end{center}
\textbf{Figure 5.2}: Evolution of Hamming distance for an organism with a high amount of introns over 6 hours. x-axis: Hamming distance, y-axis: time (in minutes)\\

As one would expect, the mutation rate is much higher and very constant, the standard derivation of the hamming distance spreads continuously.\\

\subsection{Effect of alphabet "energy"}
As explained in chapter 4.3, each alphabet has a specific "energy". To see whether the used definition of the interaction actually give a good result for robust alphabets, an evolutionary experiment has been performed.\\
 
\begin{center}
\begin{tabular}{cc}
\includegraphics[scale=0.6]{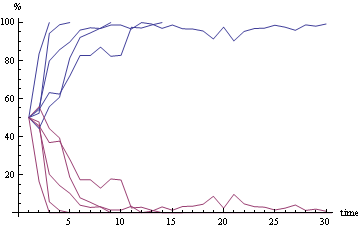} & \includegraphics[scale=0.6]{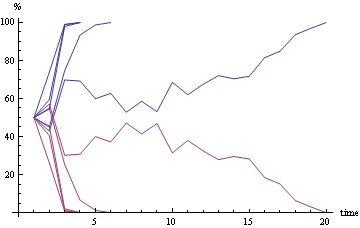}\\
\textbf{Figure 5.3}: Energy 1152.1 (blue) vs. & \textbf{Figure 5.4}: Energy 1152.1 (blue) vs. \\
1662.9 $\pm$ 11.2 (red) & 1386.8 $\pm$ 3.2 (red)
\end{tabular}
\end{center}

Figure 5.3 and 5.4 show that lower energy has a better fitness, thus the presented interaction definition is at least roughly a good approximation for an optimal alphabet.\\

\subsection{Effect of Start- and Stop-codons}
One experiment has been performed with a number of 100.000 random codons within a STOP and a START codon - a big intron. Several organisms have converted a part of the intron into an exon (by introducing a START codon) without any negative effect - all converted functional parts were neutral mutations.\\

There were two very surprising results: The biggest converted part had 33 codons, and still worked without problem. A different converted exon even managed to perform an API call without crashing.
\begin{center}
\includegraphics[scale=0.75]{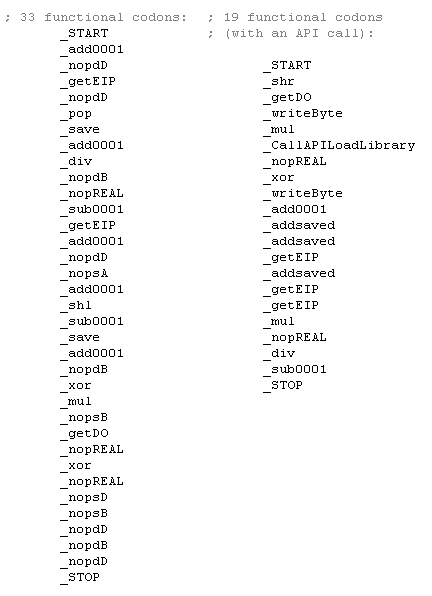}\\
\end{center}

Such huge blocks of non-lethal codons and API calls could be very valuable for artificial organisms, which are chased by behaviour scanners and API call tracers.\\

\section{Outlook}
There are several features in the natural protein biosynthesis (or in microbiology in general) which could be used in this artificial concept:\\

\textbf{Alternative Splicing:} Before translating the codons into amino acids, the splicing process cuts out introns. In natural systems, there can be an alternative splicing process, which can create the final codon sequence in many alternative ways. For example, exons can be combined in different ways, some exons could be cutted out, introns can be coded, and other methods. This process is influenced by regulatory elements (such as proteins). An analogy to this process would increase the variability of the organisms drastically.\\

\textbf{Protein Folding:} After translation of codons into amino acids, a process called \textit{folding} gives a specific 3D structure to the amino acid chain - this structure is mainly responsible for the proteins chemical properties. A similar further layer of translation may have advantages for the organisms too.\\

\textbf{Protocol for Horizontal Gene Transfer:} To exchange valuable information as antibiotic resistance, bacteria have developed processes called \textit{conjugation}. To develope a process like that for exchanging useful information within the population would be a big advantage.\\ 

There are some further techniques which would increase the variability of the organisms, such as increasing the functional code size. The difficulties come from restrictions of the PE format, which requires more than one field to be mutated in the same manner. This is a very unlikely process - finding a solution to that problem would open many new possibilities for the organisms.\\

The experiments have shown a very promising behaviour of the different mutation techniques. Especially the START- and STOP-codon experiment, where organisms performed an additional API call, indicates that even macro mutations are realistic within this concept - and behaviour changing mutations are not always lethal.\\

The question of how antivirus programs can detect organisms using this technique is open. The organism's non-lethal configuration is infinite, and due to the fact that darwinian evolution is not predictable, algorithmic approaches are probably unusable and limited. Behaviour scanners are likely (due to effects that has been shown with Splicing) not practicable, too. Statistical approaches may work for a low number of generations quite good, but will most likely fail for big difference to the ancestor (at high generations or when several macro mutations happened), as well.\cite{Mostafa1}.\\

... as a conclusion, one can see:\\
\textbf{The artificial organisms took the redpill - and enjoy their new freedom now...}

\end{document}